# TQ-HQNet: A Trainability-Aware and Qubit-Efficient Hybrid Quantum Neural Network for Cross-Domain Learning on Heterogeneous Data

Abdur Rasool
*Department of Statistics and Data Science*
*Southern University of Science and Technology*
Shenzhen, China
abdurrasool@sustech.edu.cn

Saba Aslam
*Shenzhen Institutes of Advanced Technology*
*Chinese Academy of Sciences*
Shenzhen, China
aslam@siat.ac.cn

Hui Chen
*School of Artificial Intelligence*
*Shenzhen Polytechnic University*
Shenzhen, China
huichen@szpu.edu.cn

Chayut Bunterngchit*
*Faculty of Engineering and Technology*
*King Mongkut's University of Technology, North Bangkok, Rayong Campus*
Rayong, Thailand
chayut.b@eat.kmutnb.ac.th

***Abstract*—Hybrid quantum neural networks are a promising direction for noisy intermediate-scale quantum computing, but their practical effectiveness is often limited by poor trainability, rigid circuit design, and weak robustness across heterogeneous real-world tasks. This paper proposes TQ-HQNet, a Trainability-Aware and Qubit-Efficient Hybrid Quantum Neural Network for cross-domain learning on heterogeneous data. The proposed framework combines three mechanisms: 1) adaptive latent compression to map domain-specific features into qubit-compatible representations, 2) trainability-aware search over compact variational circuit candidates, and 3) hybrid fusion of multiple shallow quantum experts to improve robustness without increasing qubit cost excessively. TQ-HQNet is evaluated on four public datasets spanning neurological screening, phishing detection, financial direction prediction, and DNA sequence classification. Across all domains, the proposed model consistently improves over a fixed hybrid quantum baseline, with balanced-accuracy gains of +0.280, +0.170, +0.071, and +0.060, respectively. Ablation and sensitivity studies confirm that gains arise from trainability-aware compact circuit selection and expert fusion, not circuit scaling. Although classical baselines outperform on some tasks, fixed-split results show consistent improvements over the fixed hybrid reference. These analyses support the practical role of compact selection and fusion, with repeated-split and hardware-aware studies left for future work.***

## I. Introduction

Quantum machine learning (QML) is widely regarded as one of the most promising application areas for noisy intermediate-scale quantum (NISQ) computing because hybrid quantum-classical optimization can be implemented with shallow parametrized circuits and classical backpropagation loops [1]–[3]. In supervised learning, prior studies have established several important foundations, including quantum feature maps [4], [5], quantum circuit learning [6], data re-uploading for expressive shallow models [7], and specialized architectures such as quantum convolutional neural networks [8]. These studies confirm that compact quantum models can learn useful decision boundaries, but they also reveal a central practical difficulty: good QML performance is highly sensitive to circuit design, optimization stability, and qubit budget. This sensitivity also depends on the data-encoding map, which can substantially affect the expressive power of variational quantum models [7].

This challenge is particularly important for heterogeneous real-world datasets [9]–[15]. Unlike synthetic benchmarks or single-domain silos, practical learning tasks often differ substantially in feature dimension, statistical structure, noise profile, and class separability. In such settings, a fixed hybrid quantum architecture may be too rigid: a circuit that is trainable and effective for one domain may become unstable or under-expressive for another. This problem is compounded by well-known trainability issues in variational quantum models, including barren plateaus and unfavorable optimization landscapes [16], as well as the broader trade-off between expressivity, compactness, and generalization [17]–[21]. Therefore, an effective QML system for heterogeneous data should not simply scale circuit depth, but should explicitly account for trainability and qubit efficiency during model design.

*Corresponding author

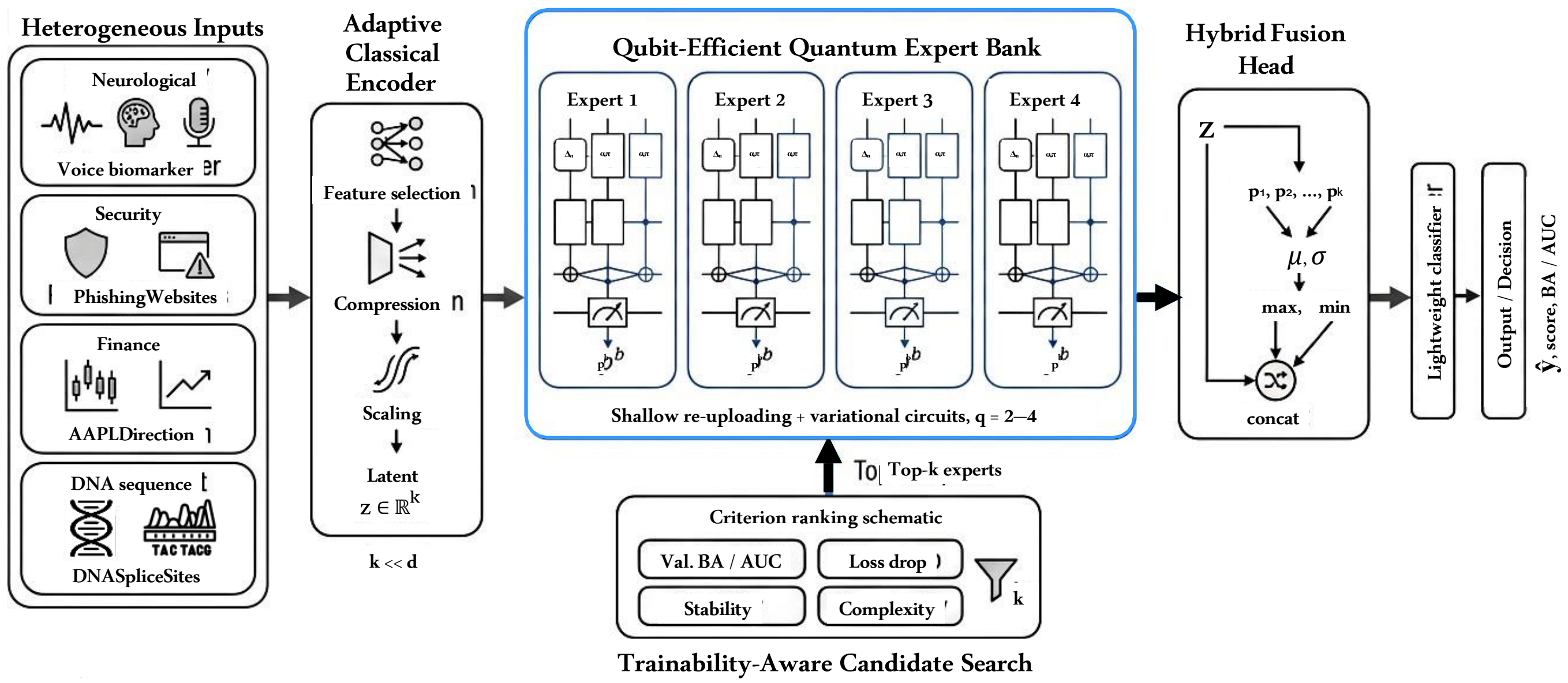

Fig. 1. Overview of the proposed TQ-HQNet architecture. The framework integrates adaptive classical compression, trainability-aware quantum expert search, and hybrid fusion to enable qubit-efficient cross-domain learning.

Motivated by this gap, this paper studies the following research question: *can a single shallow hybrid quantum architecture, when explicitly designed around trainability and qubit efficiency, outperform a fixed hybrid quantum baseline consistently across heterogeneous domains?* To answer this question, we propose ***TQ-HQNet***, a *Trainability-Aware and Qubit-Efficient Hybrid Quantum Neural Network* for cross-domain learning on heterogeneous data. The aim of this study is to develop a single hybrid QML framework that remains compact enough for NISQ settings while adapting effectively to tasks from biomedical screening, phishing detection, financial prediction, and DNA splice-junction classification.

TQ-HQNet is built around three mechanisms. First, a lightweight adaptive encoder compresses heterogeneous input features into qubit-compatible latent representations. Second, a trainability-aware search selects compact candidate circuits over qubit count, ansatz depth, and feature re-uploading depth instead of relying on a fixed architecture. Third, the selected shallow quantum experts are combined through a hybrid fusion head, following the principle that expert diversity improves robustness under heterogeneous inputs [22]. This design prioritizes stability and efficiency as first-class objectives. Our main contributions are:

- We propose TQ-HQNet, a trainability-aware and qubit-efficient hybrid QNN integrating adaptive compression, compact circuit search, and hybrid fusion.
- We formulate a trainability-aware selection strategy that scores candidates using predictive quality and optimization behavior for reliable model selection under qubit constraints.
- We validate the method on four heterogeneous public datasets, providing a cross-domain assessment of hybrid QML robustness. Through benchmark, ablation, and sensitivity analyses, we show that gains stem from trainability-aware compact circuit selection and expert fusion, not from circuit scaling alone.

Overall, this work does not claim universal superiority over classical methods. Instead, it demonstrates that trainability-aware design substantially improves the consistency, stability, and practical competitiveness of hybrid quantum learning across diverse domains under realistic NISQ constraints.

## II. Method

### A. Problem Formulation and Overview of TQ-HQNet

We consider binary classification over $\mathcal{D} = \{(x_i, y_i)\}_{i=1}^{N}$, with $x_i \in \mathbb{R}^d$ and $y_i \in \{0, 1\}$. In practice, $d \gg q$, where $q$ is the number of qubits available on near-term hardware. Standard variational quantum classifiers (VQCs) and shallow hybrid quantum neural networks (HQNNs) are sensitive to circuit architecture, encoding, and optimization dynamics, often yielding unstable training or poor generalization across domains with differing statistical structures (e.g., tabular data versus k-mer sequence representations). TQ-HQNet addresses three key challenges: (i) classical-to-quantum dimensional mismatch, (ii) trainability fragility of fixed circuits, and (iii) limited robustness of single-expert quantum classifiers on heterogeneous data. The proposed architecture is described in detail in Section I and illustrated in Fig. 1. The following subsections present the technical components of the framework.

### B. Adaptive Classical Encoder

Given a raw input $x \in \mathbb{R}^d$, an adaptive encoder $f_\vartheta$ produces a compact latent representation $z = f_\vartheta(x) \in \mathbb{R}^k$, with $k \ll d$. For tabular data, the encoder applies standardization, supervised feature selection via mutual information, and caps the latent dimension at a small value. For DNA sequence inputs, overlapping k-mer tokenization is followed by TF-IDF or count vectorization, chi-square filtering, truncated SVD, and standardization. The resulting features are scaled to $[-1, 1]$ for angle encoding, and only the first $q$ dimensions are passed to the quantum expert bank to enforce a strict qubit budget.

### C. Qubit-Efficient Quantum Expert Bank

Each quantum expert is parameterized by a candidate tuple $c = (q, r_f, r_a)$, where $q \in \{2, 3, 4\}$ is the number of qubits, $r_f$ is the feature re-uploading depth, and $r_a$ is the variational ansatz depth. Both depths are deliberately kept small to maintain trainability. The quantum state [6], [7] prepared by expert $c$ is

$$|\psi_c(z)\rangle = U_a^{(c)}(\phi_c)\, U_e^{(c)}(z)\, |0\rangle^{\otimes q}, \tag{1}$$

where $U_e^{(c)}(z)$ denotes the data-reuploading embedding circuit consisting of repeated $RY$ and $RZ$ rotations interleaved with local $CZ$ entangling gates, and $U_a^{(c)}(\phi_c)$ is a shallow trainable ansatz of $RY$, $RZ$, and $CX$ gates. The expert produces a probabilistic output [6]:

$$p_c(y = 1 \mid x) = \sigma\left(\langle \psi_c(z)|M|\psi_c(z)\rangle\right), \tag{2}$$

where $\sigma$ is the sigmoid function and $M$ is a fixed observable (typically $Z$ on the first qubit or a parity operator).

The circuit complexity [17] of a candidate is defined as

$$K(c) = q(r_f + r_a). \tag{3}$$

### D. Trainability-Aware Quantum Expert Search

Instead of committing to a single hand-designed circuit, we maintain a pool of candidate experts and evaluate each on a held-out validation split. Candidate quality is assessed by a composite trainability-aware score [16]:

$$S(c) = F\left(BA_{\text{val}}(c),\ AUC_{\text{val}}(c),\ \Delta L(c)/K(c),\ \mathrm{Std}L(c)\right), \tag{4}$$

where $BA_{\text{val}}$ and $AUC_{\text{val}}$ are validation balanced accuracy and area under the ROC curve, $\Delta L(c)$ is the observed loss reduction during optimization, and $\mathrm{Std}L(c)$ measures loss variability across training epochs.

A shortlist is first formed by retaining candidates whose validation balanced accuracy lies within a small $\epsilon$-margin of the best observed value. Among these, the final ranking prioritizes higher normalized loss drop, lower optimization variance, and lower circuit complexity. This explicit incorporation of optimization behavior (beyond raw predictive accuracy) constitutes the central methodological contribution of the trainability-aware search.

**Algorithm 1** TQ-HQNet Training Procedure

**Input:** Dataset $\mathsf{D}$, candidate pool $\mathsf{C}$, number of experts $k$
**Output:** Selected experts $\mathsf{E}^*$ and fusion head $g$

Split $\mathsf{D}$ into expert-train, fusion-train, validation and test
**for** each candidate $c \in \mathsf{C}$ **do**
  Train expert $c$ on expert-train split
  Compute Eq. (4) $S(c)$ on validation split
**end for**
Select top-$k$ experts $\mathsf{E}^*$ according to ranking on $S(c)$
Retrain selected experts $\mathsf{E}^*$ on expert-train split
Generate $h(x)$ Eq. (5) on fusion-train split using $\mathsf{E}^*$
Train logistic regression $g$ on the fusion features Eq. (6)
**return** $\mathsf{E}^*$, $g$

TABLE I
DATASETS USED IN THE CROSS-DOMAIN BENCHMARK.

| Dataset | Raw size | We Used | Representation |
|---|---|---|---|
| Parkinsons [9] | 195 | 180 | 22 acoustic voice features |
| PhishingWebsites [11] | 11,055 | 300 | 30 handcrafted website features |
| AAPLDirection [13] | 506 | 420 | 9 indicator features |
| DNASpliceSites [15] | 3,190 | 300 | 60-base DNA sequence |

### E. Hybrid Fusion Head

Rather than deploying only the single highest-ranked expert, the method retains the top-$k$ experts. Each selected expert is retrained and treated as a probabilistic feature extractor. A fusion feature vector [22] is then constructed as

$$h(x) = \left[z,\ p_1, \dots, p_k,\ \mu_p,\ \sigma_p,\ \max_p,\ \min_p\right], \tag{5}$$

where $\mu_p$, $\sigma_p$, $\max_p$, $\min_p$ are summary statistics of the expert probability outputs. A lightweight logistic regression head $g(\cdot)$ produces the final prediction [22]:

$$p(y = 1 \mid x) = g\left(h(x)\right). \tag{6}$$

By aggregating multiple complementary shallow quantum views together with the original latent classical features, the fusion head mitigates the brittleness of any individual circuit while preserving data efficiency and interpretability. The complete training procedure is summarized in Algorithm 1.

### F. Computational Intuition and Design Rationale

The framework is deliberately restricted to shallow, low-width quantum circuits, aiming not for quantum advantage through capacity, but for reliable hybrid quantum learning under NISQ constraints. Adaptive compression resolves the dimensional mismatch between high-dimensional inputs and tiny quantum circuits. Trainability-aware search avoids circuits that appear strong on validation accuracy yet exhibit unstable optimization. Multi-expert fusion replaces the single point of failure in conventional single-circuit HQNNs with a robust aggregated decision mechanism. Together, these choices make hybrid quantum learning substantially more practical and stable across heterogeneous domains.

TABLE II
MAIN HYPERPARAMETERS AND EXPERIMENTAL SETTINGS.

| Component | Setting |
|---|---|
| Tabular preprocessing | StandardScaler → SelectKBest(mutual information) → MinMaxScaler([−1, 1]) |
| Sequence preprocessing | TF–IDF (max_features=1200, 1–2 grams, min_df=2) → SelectKBest($\chi^2$) → TruncatedSVD → StandardScaler → MinMaxScaler([−1, 1]) |
| Latent dimension $d_z$ | 8 for Parkinsons/PhishingWebsites/AAPLDirection; 6 for DNASplice |
| Fixed-HQNet | $q$=4, $r_f$=2, $r_a$=2 for Parkinsons/PhishingWebsites/AAPLDirection; $q$=4, $r_f$=1, $r_a$=1 for DNASplice |
| Quantum search space | Parkinsons, PhishingWebsites: $q \in \{2, 3, 4\}$, $r_f \in \{1, 2\}$, $r_a \in \{1, 2\}$; AAPLDirection: $q \in \{3, 4\}$, $r_f \in \{1\}$, $r_a \in \{1, 2\}$; DNASplice: $q$=4, $r_f$ =1, $r_a$=1 |
| Search iterations | 5, 5, 4, 3 (Parkinsons, PhishingWebsites, AAPLDirection, DNASplice) |
| Search restarts | 2, 2, 1, 1 |
| Final/expert iterations | 10, 10, 8, 6 |
| Top-$k$ experts | 3, 3, 2, 1 |
| Expert restarts | 2, 2, 1, 1 |
| Fusion head | Logistic regression, $C = 0.8$, class_weight=balanced, max_iter=1200 |
| Classical baselines | LogReg (max_iter=1000), RBF-SVM (probability=True), MLP (hidden size 16, max_iter=500) |

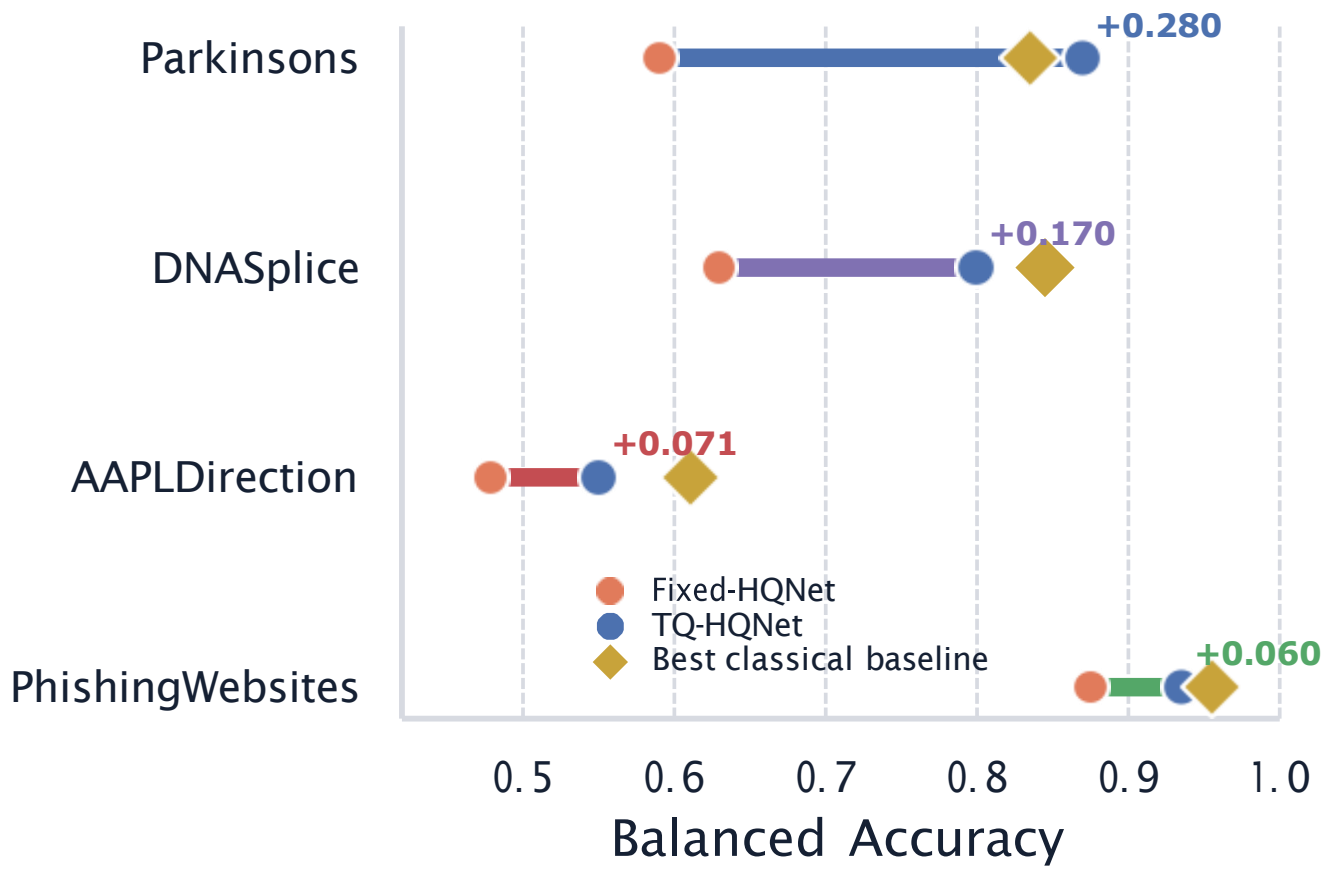

Fig. 2. Cross-domain benchmark in balanced accuracy. TQ-HQNet consistently outperforms Fixed-HQNet across all datasets, with the largest gains on Parkinsons and DNASplice.

## III. EXPERIMENTS AND RESULTS

### A. Experimental Settings

*a) Datasets and Preprocessing:* We evaluated TQ-HQNet on four heterogeneous public datasets covering neurological disorder diagnosis [9], phishing detection [11], financial direction classification [13], and DNA splice-junction classification (binary splice-site vs. non-splice) [15] (Table I). DNASpliceSites is a binary formulation of the public UCI splice-junction dataset, where EI and IE are merged into a splice-junction class, and N is treated as the non-splice class. For each dataset, we first applied stratified subsampling to keep the quantum experiments computationally feasible, and then split the data into train/validation/test partitions with a 70/15/15 ratio using a fixed random seed of 7. All preprocessing operators were fit on the training split only and then applied to the validation and test splits.

For tabular datasets, we used standardization followed by mutual-information-based feature selection to a Latent dimension $d_z$, and finally scaled the selected features to $[-1, 1]$ for angle encoding. For DNA sequence datasets, we used k-mer TF–IDF + chi–square with unigrams and bigrams, selected the top features with $\chi^2$, reduced them with truncated singular value decomposition (SVD) to $d$, and again rescaled the latent vectors to $[-1, 1]$. Cross-domain denotes evaluation across separate heterogeneous datasets; no transfer learning or domain adaptation is performed.

*b) Evaluation Metrics and Hyperparameters:* We report test Accuracy, Balanced Accuracy, F1-score, and AUC, while the validation split was used only for trainability-aware architecture search. The main hyperparameters and experimental settings are summarized in Table II.

*c) Quantum Search Space and Training:* The quantum search space is defined by the number of qubits $q$, feature re-uploading depth $r_f$, and ansatz depth $r_a$, with the complexity proxy $q(r_f + r_a)$. For each candidate, we trained a variational quantum classifier (VQC) with COBYLA and retained the best restart according to validation balanced accuracy, validation AUC, normalized loss reduction, and loss stability. The final single-expert architecture was selected from candidates within $\epsilon = 0.02$ of the best validation balanced accuracy, preferring larger normalized loss reduction, higher validation F1, and lower complexity.

*d) Expert Fusion Setup:* For TQ-HQNet, we ranked the searched candidates and retained the top-$k$ unique experts. The training split was further divided into an expert-training subset and a fusion-head subset (75/25). Each quantum expert produced probabilistic outputs, and the final classical fusion head was trained on the concatenation of the latent input, expert probabilities, and summary statistics (mean, standard deviation, maximum, and minimum). This design lets the model exploit complementary shallow quantum experts without increasing the qubit budget.

*e) Baselines and Ablations:* We compared the proposed model against logistic regression (LogReg), RBF-kernel SVM (RBF-SVM), and a one-hidden-layer MLP. We also compared against *Fixed-HQNet*, a manually fixed single quantum classifier without trainability-aware search or expert fusion. To analyze the source of gains, we considered four ablation variants: A0 (*FixedSingle*), a fixed single VQC; A1 (*AdaptiveSingle*), a trainability-aware selected single VQC; A2 (*FixedFusion*), a fixed quantum expert bank with hybrid fusion; and A3 (*AdaptiveFusion*), the TQ-HQNet model. We further evaluated sensitivity to the number of selected experts $k$ and expert restarts.

### B. Main Cross-Domain Benchmark

As shown in Fig. 2 and Table III, TQ-HQNet consistently improves over the non-adaptive quantum baseline across all four domains, showing consistent improvement over the fixed hybrid reference in four separate within-dataset evaluations. In balanced accuracy, TQ-HQNet improves over Fixed-HQNet

TABLE III
CROSS-DOMAIN BENCHMARK PERFORMANCE. TQ-HQNET CONSISTENTLY OUTPERFORMS FIXED-HQNET ACROSS ALL DOMAINS AND ACHIEVES THE BEST QUANTUM PERFORMANCE.

| Dataset | Model | Acc | Bal. Acc | F1 | AUC |
|---|---|---|---|---|---|
| Parkinsons [9] | LogReg | 0.9000 | 0.8354 | 0.9362 | 0.9255 |
| | RBF-SVM | **0.9000** | 0.8354 | **0.9362** | **0.9814** |
| | MLP | **0.9000** | 0.8354 | **0.9362** | **0.9814** |
| | Fixed-HQNet | 0.6000 | 0.5901 | 0.7000 | 0.6646 |
| | TQ-HQNet | 0.8000 | **0.8696** | 0.8500 | 0.9441 |
| Phishing [11] | LogReg | 0.8889 | 0.8850 | 0.9020 | 0.9720 |
| | RBF-SVM | **0.9556** | **0.9550** | **0.9600** | **0.9920** |
| | MLP | 0.9111 | 0.9100 | 0.9200 | 0.9800 |
| | Fixed-HQNet | 0.8889 | 0.8750 | 0.9091 | 0.8650 |
| | TQ-HQNet | 0.9333 | 0.9350 | 0.9388 | 0.9820 |
| Finance [13] | LogReg | **0.6349** | **0.6107** | **0.7160** | **0.5867** |
| | RBF-SVM | 0.5079 | 0.4857 | 0.6076 | 0.5306 |
| | MLP | 0.5079 | 0.4929 | 0.5867 | 0.4673 |
| | Fixed-HQNet | 0.5238 | 0.4786 | 0.6739 | 0.3949 |
| | TQ-HQNet | 0.5556 | 0.5500 | 0.6000 | 0.5276 |
| DNA Seq. [15] | LogReg | 0.8000 | 0.7984 | 0.8163 | 0.9289 |
| | RBF-SVM | **0.8444** | **0.8449** | **0.8444** | **0.9348** |
| | MLP | 0.8000 | 0.7994 | 0.8085 | 0.9308 |
| | Fixed-HQNet | 0.6222 | 0.6294 | 0.4516 | 0.7411 |
| | TQ-HQNet | 0.8000 | 0.7994 | 0.8085 | 0.9209 |

Acc = Accuracy, Bal. Acc = Balanced Accuracy. Best results are in **bold**, second-best are underlined.

by +0.280 on Parkinsons, +0.170 on DNASplice, +0.071 on AAPLDirection, and +0.060 on PhishingWebsites, with an average gain of +0.145. The strongest gain is on Parkinsons, where the model also surpasses the best classical baseline in balanced accuracy (0.8696 vs. 0.8354), indicating that hybrid fusion is especially beneficial for compact biomedical descriptors under limited budgets. On PhishingWebsites, the method remains highly competitive with a balanced accuracy of 0.9350 and AUC of 0.9820, showing effectiveness on high-separability security data. Although classical baselines remain stronger on AAPLDirection and DNASplice, the proposed model delivers consistent gains over the fixed quantum architecture, which is the central technical claim of this work.

### C. Architectural Ablation Study

Fig. 3(a) shows that each component of TQ-HQNet contributes meaningfully. Starting from A0, adding adaptive search alone (A1) increases mean balanced accuracy by +0.090, showing that trainability-aware selection removes a substantial portion of optimization brittleness. Hybrid fusion (A2) yields the largest improvement, with a mean gain of +0.202 over A0, indicating that expert-level diversity and aggregation are the dominant factors behind cross-domain robustness. The full model A3 retains a strong +0.193 gain over A0, matching the best ablation on Parkinsons and PhishingWebsites. As shown in Fig. 3(b), the small decrease of A3 relative to A2 on AAPLDirection and DNASplice suggests that trainability-aware selection slightly sacrifices peak dataset-specific performance for a more stable and qubit-efficient configuration, consistent with avoiding brittle high-variance circuit choices.

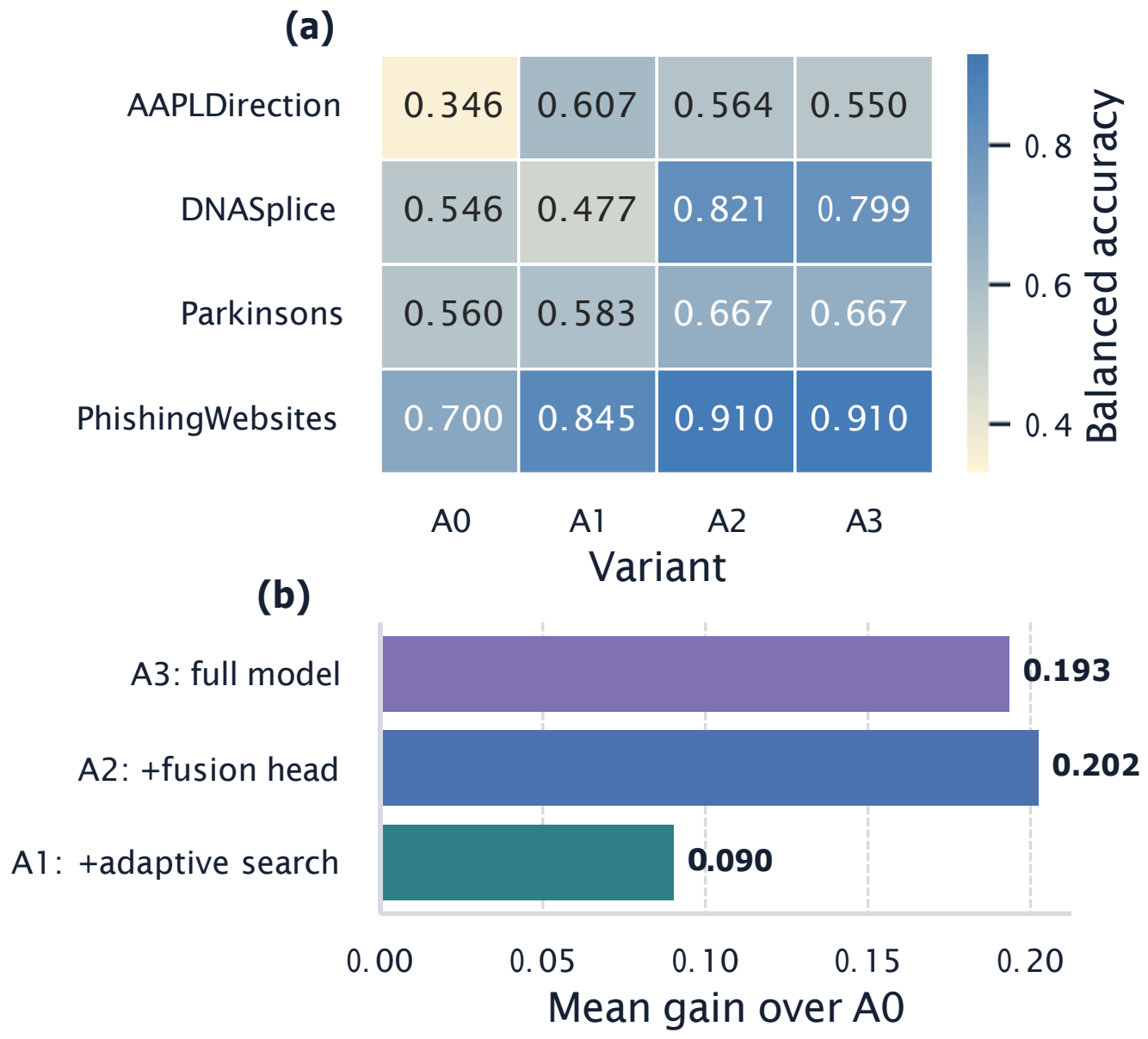

Fig. 3. Architectural ablation study. (a) Balanced accuracy of four variants: A0 (fixed single-expert baseline), A1 (+adaptive search), A2 (+hybrid fusion), and A3 (full model). (b) Mean accuracy gain over A0, showing hybrid fusion and adaptive selection as the main drivers of improvement.

### D. Trainability-Efficiency Trade-off and Expert Sensitivity

Fig 4 provides further evidence for why TQ-HQNet remains effective under strict qubit budgets. In Fig. 4(a), each point denotes one searched circuit candidate, where the horizontal axis uses the complexity proxy $q(r_f + r_a)$, with $q$ the number of qubits, $r_f$ the feature re-uploading depth, and $r_a$ the ansatz depth. The annotations mark the selected candidates for each domain, with PD, PHI, FIN, and DNA denoting Parkinsons, PhishingWebsites, AAPLDirection, and DNASplice, respectively; for example, "PD $q$=4" indicates a Parkinsons candidate using four qubits. The selected models cluster in the low-to-moderate complexity region, showing that better validation behavior is obtained not by maximizing circuit size, but by choosing compact 3- to 4-qubit configurations with more favorable trainability-efficiency trade-offs. Fig. 4(b) then analyzes sensitivity to the number of selected experts ($k$) and the number of expert restarts. Performance is largely stable across these settings for Parkinsons, AAPLDirection, and DNASplice, while PhishingWebsites shows only a small gain from the weakest setting to the stronger ones. This saturation trend indicates that the proposed model does not rely on a large expert pool or repeated restarts to remain competitive, which directly supports the qubit-efficient design of TQ-HQNet.

## IV. CONCLUSION

This work shows that trainability-aware circuit selection and hybrid expert fusion can improve the observed balanced accuracy of a compact hybrid quantum classifier relative to a fixed single-circuit reference on four separate heterogeneous tasks. The results are fixed-split point estimates, and they do not establish quantum advantage or transfer across domains. Future work includes exploring stronger quantum feature encoding strategies, more advanced optimization techniques, and validation on larger-scale datasets to further bridge the gap with classical baselines.

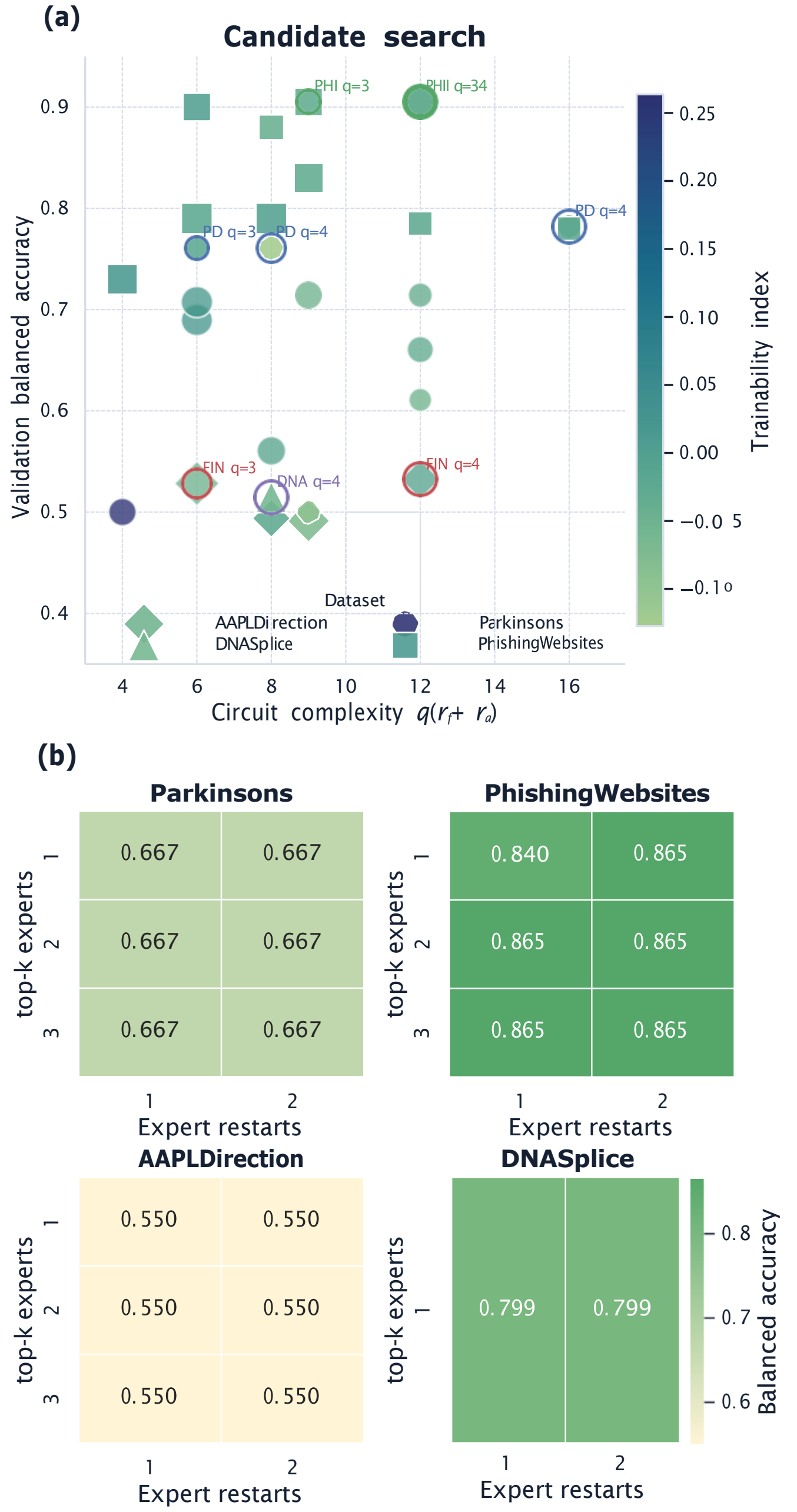

Fig. 4. Trainability-efficiency trade-off and expert-count sensitivity. (a) Candidate-search landscape over complexity $q(r_f + r_a)$, where $q$ is the number of qubits, $r_f$ is re-uploading depth, and $r_a$ is ansatz depth; selected candidates are annotated for Parkinsons (PD), PhishingWebsites (PHI), AAPLDirection (FIN), and DNASplice (DNA). (b) Sensitivity of balanced accuracy to selected experts ($k$) and restarts, with performance stable across most settings.

## Code and Datasets

The source code is available at https://github.com/abdul-rasool/TQ-HQNet-Quantum-ML, and the datasets are described in Section III-A0a.